\documentclass[conference]{IEEEtran}
\IEEEoverridecommandlockouts
\usepackage{cite}
\usepackage{amsmath,amssymb,amsfonts}
\usepackage{algorithmic}
\usepackage{graphicx}
\usepackage{textcomp}
\usepackage{xcolor}
\def\BibTeX{{\rm B\kern-.05em{\sc i\kern-.025em b}\kern-.08em
    T\kern-.1667em\lower.7ex\hbox{E}\kern-.125emX}}
\begin{document}

\title{Periocular Recognition Using CNN Features Off-the-Shelf
\thanks{Authors thank the Swedish Research Council (VR), the Sweden's innovation agency (VINNOVA),
and the Swedish Knowledge Foundation (CAISR program and SIDUS-AIR project).}
}

\author{\IEEEauthorblockN{Kevin Hernandez-Diaz, Fernando Alonso-Fernandez and Josef Bigun}
\IEEEauthorblockA{\textit{School of Information Technology (ITE)},
\textit{Halmstad University}, Box 823, 30118 Halmstad, Sweden \\ Email: kevin.hernandez-diaz@hh.se, feralo@hh.se, josef.bigun@hh.se}
}

\maketitle

\begin{abstract}
Periocular refers to the region around the eye, including sclera, eyelids, lashes, brows and skin.
With a surprisingly high discrimination ability, it is the ocular modality requiring the least constrained acquisition.
Here, we apply existing pre-trained architectures, proposed in the context of the
ImageNet Large Scale Visual Recognition Challenge, to the task of periocular recognition.
These have proven to be very successful for
many other computer vision tasks apart from the detection and classification tasks for which they were
designed.
Experiments are done with a database of periocular images captured with a digital camera.
We demonstrate that these off-the-shelf CNN features can effectively recognize individuals based on
periocular images, despite being trained to classify generic objects.
Compared against reference periocular features, they show an EER reduction of up to $\sim$40\%,
with the fusion of CNN and traditional features providing additional improvements.
\end{abstract}


\section{Introduction}

Periocular biometrics represent a trade-off between using
the iris (which may not be available in long distances due to low resolution, e.g. surveillance)
and the face (which may be partially occluded, either accidentally or intentionally).
It has attracted noticeable interest in recent years,
being one the most discriminative regions of the face \cite{[Alonso16]}.
It is also very suitable for non-cooperative biometrics, since it can
be captured largely relaxing the acquisition conditions. 
%
In addition, it has shown to be more resistant to expression variation \cite{[Smereka15]},
aging \cite{[Juefei-Xu11]},
plastic surgery \cite{[Jillela12]},
or gender transformation \cite{[Mahalingam14]},
as compared with the entire face.
Apart from serving as a stand-alone modality, it can also be combined with
iris or face \cite{[Alonso15a]}
to improve the overall performance.

Convolutional Neural Networks (CNNs) have become a very popular tool in many vision tasks.
Their application to biometrics is however limited,
with recent works on face recognition \cite{[Parkhi15]}
and detection \cite{[Li15]},
%
iris recognition \cite{[Nguyen18]},
soft-biometrics,
and image segmentation 
\cite{[Bhanu17bookDLbio]}.
One reason of such limited research is the lack of big amounts of
training data, as required by deep learning methods.
Inspired by the work of Nguyen \emph{et al.} in iris \cite{[Nguyen18]},
this paper leverages the power of existing pre-trained architectures which have proven to
be successful in very large recognition tasks.
This eliminates the necessity of designing and training new CNNs for the specific task
of periocular recognition, which may be infeasible given the mentioned lack of
large-scale data.
In particular, we choose those from the series
of the ImageNet Large Scale Visual Recognition Challenge (ILSVRC).
%
These are trained on more than a million images of the ImageNet database\footnote{http://www.image-net.org},
and it can classify images into 1000 object categories.
Apart from obtaining leading positions in ILSVRC, these off-the-shelf networks have also proven
very successful in many other tasks apart from the object detection and classification
for which they are designed 
\cite{[Razavian14]}.
Here, we investigate those architectures
in the context of periocular recognition.
We compare them against traditional baseline Local Binary
Patterns (LBP) \cite{[Ojala02]}, Histogram of Oriented Gradients
(HOG) \cite{[Dalal05]} and Scale-Invariant Feature Transform (SIFT)
key-points \cite{[Lowe04]}. We employ a database of 1,718 periocular images captured
with a digital camera at several distances.
Our experiments show that Off-the-Shelf CNN features from these popular architectures
can outperform traditional features, with Equal Error Rate (EER) reductions of up to $\sim$40\%.
We also show that fusion of CNN and traditional features provides additional
performance improvements.


\begin{figure*}[t]
\centering
\includegraphics[width=0.7\textwidth]{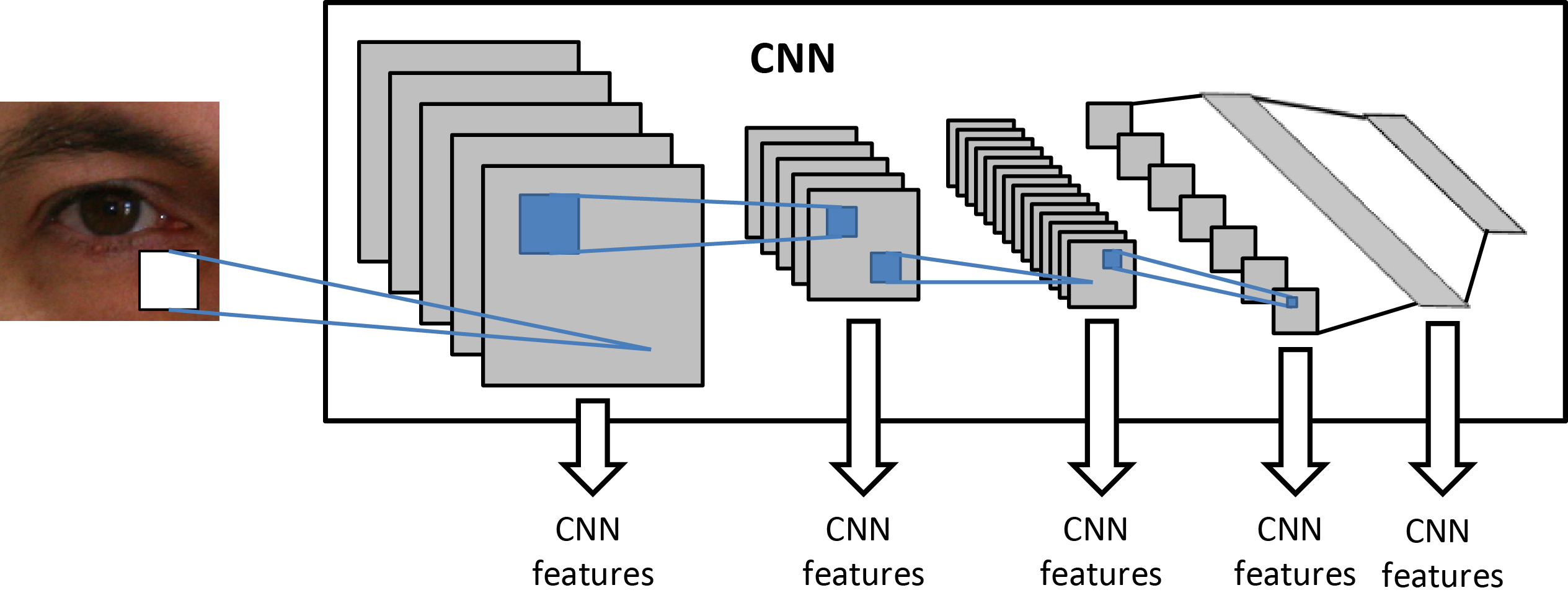}
\caption{Extraction of periocular features from different CNN layers.} \label{fig:cnns}
\end{figure*}

\section{Periocular Recognition Using CNN Features}
\label{sec:CNNs}

This section describes the basics of the CNN architectures used.
Features are extracted from the different layers (Figure~\ref{fig:cnns}).
Apart from convolutional layers, we also extract features from
max pooling, ReLU, dropout, and fully connected layers.
For this purpose, we employ the pre-trained models in Matlab r2018a.

\noindent \textbf{AlexNet} \cite{[Krizhevsky12]}
obtained the 1$^{st}$ position in ILSVRC 2012. 
The network achieved a breakthrough in this competition, with a top-5 error
of 15.3\%, more than 10.8\% percentage points ahead of the runner up.
It has 25 layers (including 5 convolutional layers).

\begin{figure}[b]
\centering
\includegraphics[width=0.48\textwidth]{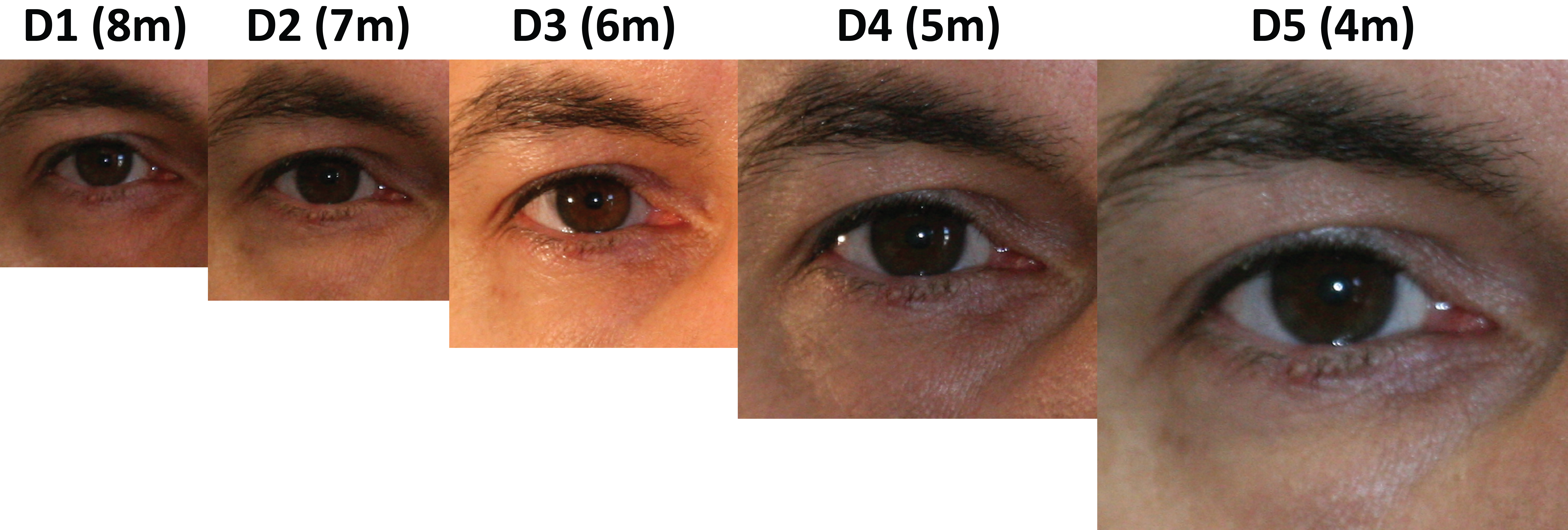}
\caption{Images from the UBIPr database (relative scale difference between images is shown).} \label{fig:db-samples}
\end{figure}

\noindent \textbf{GoogLeNet/Inception v1} \cite{[Szegedy15]}
is the winner of ILSVRC 2014 with a top-5 error of 6.7\%.
Its novelty is the use of inception modules.
These employ convolution filters of different sizes in the same layer,
allowing multi-level feature extraction.
To reduce the number of parameters, convolutions of 1$\times$1
are first applied. 
%
%
Improvements were added in later versions
by redesigning the filter arrangement in the inception module,
and by employing more inception modules \cite{[Szegedy16],[Szegedy17]}.
%
Here, we use GoogLeNet/Inception v1, with 144 layers
(including 22 convolutional layers).

\noindent \textbf{ResNet} \cite{[He16]}
introduced the concept of
residual connections to ease the training of CNNs. By reducing
the number of training parameters, they can be substantially deeper.
The key idea is to make available the input of a lower layer
to a higher layer, bypassing intermediate ones.
With an ensemble of residual networks,
they won ILSVRC 2015, with a top-5 error of 3.57\%.
Here, we employ ResNet50 and ResNet101, having 177 and 347 layers
(including 50 and 101 convolutional layers respectively).

\noindent \textbf{VGG and VGG-Face.}
Runner-up of ILSVRC 2014, 
VGG \cite{[Simonyan14]} 
is a deeper network (in comparison to AlexNet in 2012)
in which complexity is kept tractable by using
very small 3$\times$3 convolutional filters sequentially
that emulate larger ones.
The most popular versions are VGG16
and VGG19 (with 16 and 19 convolutional layers respectively).
Here, we employ VGG16, which has 41 layers in total.
They later presented VGG-Face \cite{[Parkhi15]}, based on VGG16,
evaluated with $\sim$1 million images from the Labeled Faces in the Wild \cite{[Huang07]}
and YouTube Faces \cite{[Wolf11]} datasets.

\begin{figure*}[t]
     \centering
     \includegraphics[width=.9\textwidth]{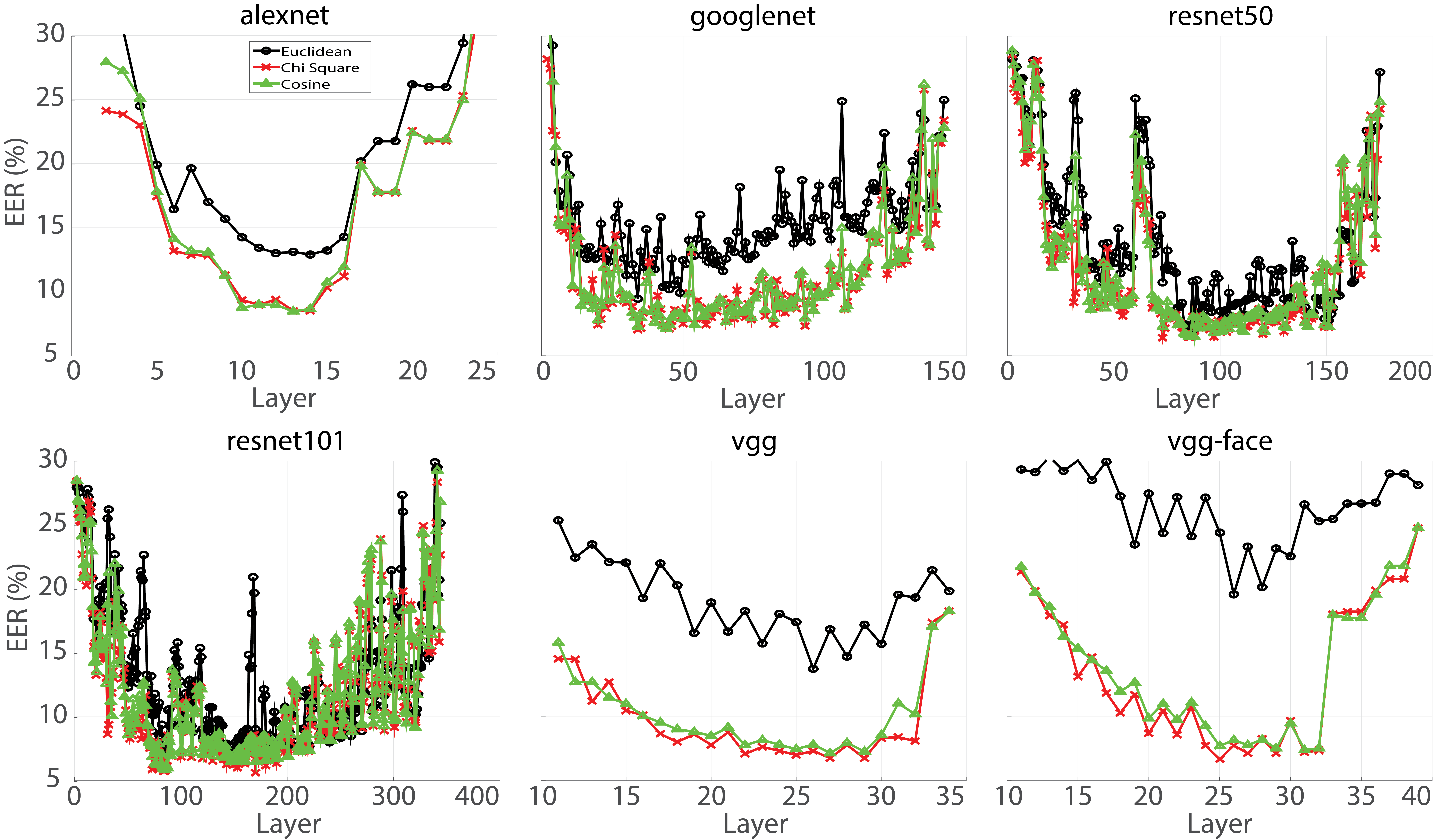}
     \caption{Verification results (EER) of different CNN layers.}
     \label{fig:EERnets}
\end{figure*}

\section{Experimental Framework}
\label{sec:exp-frame}

\subsection{Baseline Systems}

As baseline descriptors, we employ the most widely used features in
periocular research \cite{[Alonso16]}: LBP, 
HOG, 
and SIFT. 
%
In \textbf{HOG} and \textbf{LBP}, the image is decomposed
into 8$\times$8 non-overlapped 
regions.
%
%
Then, HOG and LBP features are extracted from each block. In HOG,
the histogram of
gradient orientations is built, with each bin accumulating corresponding
gradient magnitudes. In LBP, each pixel is assigned a 8-bits label
by thresholding the intensity with each pixel in the
3$\times$3 neighborhood. The binary labels 
are then
converted into decimal values and accumulated into an histogram.
Both HOG and LBP are quantized into 
8 bins histogram per block. Histograms from each block are then
normalized to account for local illumination and contrast
variations, and concatenated into a single descriptor of
the whole image.
%
%
Authentication with HOG and LBP is done by simple distance
measures. Euclidean distance is usually used \cite{[Park11]},
but it has been observed that the $\chi^2$ distance gives
better results with normalized histograms,
therefore we will use it in this work.
Regarding the \textbf{SIFT} descriptor, key-points extraction is
done first using difference of Gaussians 
functions in the
scale space.
A 
feature vector of 
$4 \times 4 \times 8 = 128$
is then obtained by computing 8-bin gradient orientation histograms
(relative to the dominant orientation to achieve rotation
invariance) in $4 \times 4$ sub-regions around the key-point.
The recognition metric is the number of paired key-points between
two images. 
%
We employ a free implementation of the SIFT
algorithm\footnote{http://vision.ucla.edu/ vedaldi/code/sift/assets/sift/index.html},
with the adaptations described in \cite{[Alonso09]}.
Particularly, it includes a post-processing step
to remove spurious pairings by imposing additional constraints to
the angle and distance of paired key-points.

\subsection{Database and Protocol}

We employ the UBIPr periocular database \cite{[Padole12]}.
It was acquired with a CANON EOS 5D digital camera in 2
sessions, with distance, illumination, pose and occlusion
variability. The distance varies between 4-8m in steps of 1m, with
resolution from 501$\times$401 pixels (8m) to 1001$\times$801 (4m).
For our experiments, we select 1,718 frontal-view images from 86
individuals (corresponding to users with 2 sessions), having
86$\times$2=172 available eyes. Two images are available per eye and
per distance, resulting in 172$\times$2=344 images per distance.
%
All images have been annotated manually, so radius and center of
the pupil and sclera circles are available. 
%
Images of each distance group are resized via bicubic interpolation
to have the same sclera radius (we choose the
average sclera radius $R_s$ of each distance group, given by the
ground truth). We use the sclera for normalization since it is
not affected by dilation (as it is the case of the pupil).
Then, images are aligned by extracting a square region of $7.6 R_s
\times 7.6 R_s$ around the sclera center. This size is set
empirically to ensure that all images have sufficient
margin to the four sides. 
Further, images are converted to gray-scale, and contrast is
enhanced by Contrast-Limited Adaptive Histogram Equalization
(CLAHE) \cite{[Zuiderveld94clahe]} to compensate variability in
local illumination. We employ CLAHE since it is usually the
preprocessing choice with images of the eye region
\cite{[Rathgeb10]}.
%
An example of normalized images is shown in Figure
\ref{fig:db-samples}.
Additionally, we position a mask on the whole iris region, in line
with \cite{[Park11]}. This is to ensure that periocular recognition
is not achieved by using information from captured high quality iris
patterns 
\cite{[sequeira17crosseyed]}.
Finally, we resized the images to match the input size
of each CNN, and subtract the pixel-wise mean image
of the whole database to improve recognition.

%
%
%
%
%
%
%
%

We carry out verification experiments, with each eye considered a
different user.
For genuine trials, we compare all
images of a user among them, avoiding symmetric comparisons. This
results in 7,722 user scores.
Concerning impostor experiments, the first image of a user is used
as enrolment image, and it is compared with the second image of the
remaining users, resulting in 172$\times$171=29,412 scores.
%
Fusion experiments are also done between different comparators. We
use linear logistic regression fusion. Given $N$ comparators which
output the scores ($s_{1j}, s_{2j}, ... s_{Nj}$) for an input trial
$j$, a linear fusion of these scores is: $f_j = a_0 + a_1 \cdot
s_{1j} + a_2 \cdot s_{2j} + ... + a_N \cdot s_{Nj}$. The weights
$a_{0}, a_{1}, ... a_{N}$ are trained via logistic regression
\cite{[Alonso08]} using two-fold cross validation.
We use this trained fusion approach because it
has shown better performance than simple fusion rules (like the mean
or the sum rule) \cite{[Alonso08]}.
Nonetheless, this is a weighted sum rule, though the coefficients
are optimized by a specific rule \cite{[Bigun97]}.

\begin{table*}[htb]
\begin{center}
\begin{tabular}{ccccccccccc}

 \multicolumn{3}{c}{baseline} &  \multicolumn{1}{c}{} &  \multicolumn{3}{c}{CNNs} &  \multicolumn{1}{c}{} &  \multicolumn{3}{c}{baseline+CNNs} \\ \cline{1-3} \cline{5-7} \cline{9-11}


\multicolumn{1}{c}{\textbf{system}} &
 \multicolumn{1}{c}{\textbf{EER}} &
 \multicolumn{1}{c}{\textbf{FRR}} &
  \multicolumn{1}{c}{\textbf{}} &
\multicolumn{1}{c}{\textbf{system}} &
 \multicolumn{1}{c}{\textbf{EER}} &
 \multicolumn{1}{c}{\textbf{FRR}} &
  \multicolumn{1}{c}{\textbf{}} &
\multicolumn{1}{c}{\textbf{system}} &
 \multicolumn{1}{c}{\textbf{EER}} &
 \multicolumn{1}{c}{\textbf{FRR}}\\ \cline{1-3} \cline{5-7} \cline{9-11}

LBP & 17.8\% & 48.0\% & & alexnet & 8.4\% & 26.1\% &  & alexnet & 6.9\% & 17.5\% \\

HOG & 11.3\% & 31.2\% & &  googlenet & 7.0\% & 15.9\% &  & googlenet & 6.0\% & 13.3\% \\

SIFT & 16.6\% & 30.1\% & &  resnet50 & 6.4\% & 15.4\% &  & resnet50 & 5.8\% & 12.7\% \\

all & 9.1\% & 21.8\% & &  resnet101 & \textbf{5.6\%} & \textbf{14\%} &  & resnet101 & \textbf{5.1\%} & \textbf{11.3\%} \\

all$^*$  & 16.0\% & n/a & &  vgg16 & 6.8\% & 16.6\% &  & vgg16 & 6.1\% & 13.2\% \\

\multicolumn{3}{l}{$^*$According to \cite{[Padole12]}} & & vgg-face & 6.7\% & 17.9\% &  & vgg-face & 5.5\% & 13.5\% \\ \cline{1-3} \cline{5-7} \cline{9-11}

%
%
%
%
%


\end{tabular}

\end{center}
\caption{Verification results of the baseline systems (left), of the
best layer of each CNN (center), and of the fusion experiments (right). FRR is given at FAR=1\%.}
\label{tab:verification-results}
\end{table*}
\normalsize

%

\begin{figure*}[t]
     \centering
     \includegraphics[width=.9\textwidth]{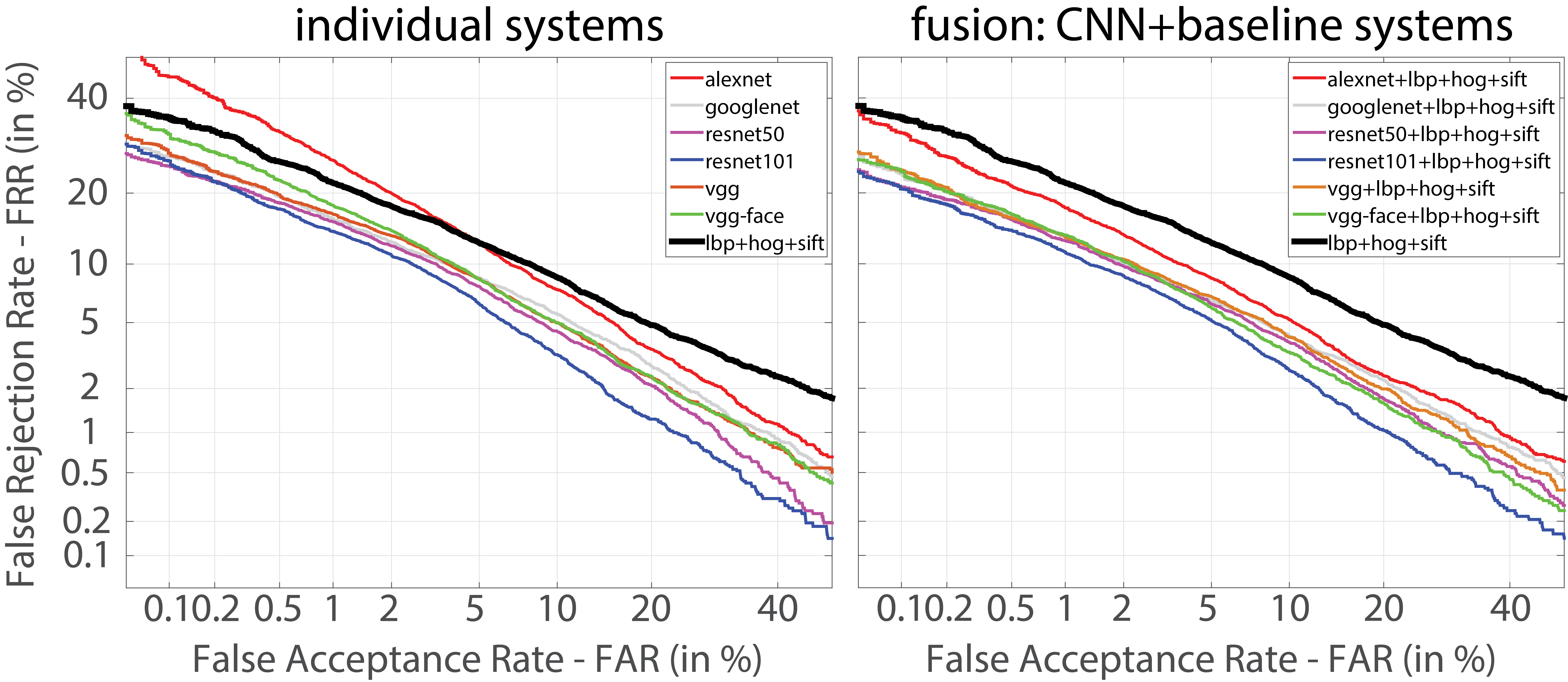}
     \caption{Verification results (DET curves). Left: individual systems
     (baseline systems and best layer of each CNN). Right: fusion experiments
     (CNN+baseline systems).}
     \label{fig:DETs}
\end{figure*}


\subsection{Results}

Normalized periocular images are fed into the feature extraction
pipeline of each CNN.
To investigate the representation capability
of each layer, 
we report the corresponding recognition accuracy using features from each layer.
%
%
The extracted CNN feature vectors are compared by simple distance or
similarity measures. In particular, we evaluate the
Euclidean distance, the $\chi^2$ distance, and the cosine similarity.
The EER is given in Figure~\ref{fig:EERnets}. 
As it can be observed, the $\chi^2$ distance and the cosine similarity consistently provide
better results than the Euclidean distance.
This is an indication that the norms of the feature vectors are not informative,
because cosine similarity and Euclidean distance are mathematically equivalent if vectors have the (Euclidean) norm 1.
It is also worth noting that EER 
has the minimum
in some intermediate layers for all CNNs.
The lowest EER is obtained at (refer to Figure~\ref{fig:EERnets}):
layer 13 with cosine similarity (AlexNet),
layer 34 with $\chi^2$ distance (GoogLeNet),
layer 73 with $\chi^2$ distance (ResNet50),
layer 170 with $\chi^2$ distance (ResNet101),
layer 27 with $\chi^2$ distance (VGG16),
and layer 25 with $\chi^2$ distance (VGG-Face).
%
%
The more intricate architectures of
GoogLeNet (with inception layers which act as small networks inside a larger network)
and Resnet (which skips connections to bypass nodes)
may explain the oscillations observed between consecutive layers.
The simpler architecture of AlexNet and VGG (which consists of single layers
stacked one after the other) results in a smoother behavior.
Also, according to \cite{[Nguyen18]},
the inception modules in GoogLeNet are able to quickly converge to the minimum,
as can be seen in Figure~\ref{fig:EERnets}. 
ResNet, on the other hand, results in a minimum at a deeper depth (specially ResNet50)
due to skipping connections between layers.
AlexNet and VGG also converge at deeper depths.
AlexNet has its minimum around layer 13, which in the Matlab implementation employed
is after the 4$^{th}$ convolutional layer (out of 5 convolutional layers).
VGG networks have the minimum in layers 25-27, which is after the
11$^{th}$ convolutional layer (out of 16 convolutional layers).
Since VGG filters are smaller, it is expected that it takes more convolutional layers
than other networks
until the optimum performance is achieved.

In the remainder of this paper, we employ the best configuration of each CNN
mentioned above. 
DET curves and performance figures of those cases
are shown in Figure~\ref{fig:DETs} (left) and
Table~\ref{tab:verification-results} (center).
We also provide the performance of
LBP, HOG and SIFT descriptors,
and the EER of the SIFT+LBP+HOG combination from a
previous study using the same database with only frontal images
\cite{[Padole12]}. It should be noted that another setup of the
matchers is used in that study which may explain the different results,
including different ROI
selection, a different fusion scheme and the use of Euclidean
distance with LBP and HOG.
%
As it can be observed, the CNNs outperform the EER of the baseline
features in all cases, as well as the FRR in most cases.
The best results are given by the ResNet architectures,
which also happens to be the deepest and the best performing ones at the
ILSVRC challenges.
The best CNN is ResNet101, with an EER of 5.6\% and
a FRR of 14\% at FAR=1\%. This is about 40\% less than the performance of
the baseline features.
GoogLeNet and VGG also provide good FRR performance.
AlexNet, on the contrary, is the worst performing CNN, which may be attributed
to its simpler and shallower structure.
It is also interesting to note that VGG-Face does not perform as well as VGG
or other CNNs, despite being tuned to work with face images.

To study complementarity between CNN and baseline features, we also
carry out fusion experiments, with results given in Figure~\ref{fig:DETs} (right)
and Table~\ref{tab:verification-results} (right).
The performance in all cases is further improved w.r.t. the individual systems.
The best combination involves (again) ResNet101, with an EER of 5.1\% and
a FRR of 11.3\%. 
This is only a 8\% improvement in the EER,
but nearly a 20\% improvement in the FRR. This is interesting for
high security applications, since operating points are usually defined at small
FAR values.


%
%
%
%
%
%
%
%
%
%
%
%
%
%

\section{Conclusions}
\label{sec:conclusions}

This paper deals with the task of periocular recognition using off-the-shelf
CNN architectures, pre-trained in the context of the
ImageNet Large Scale Visual Recognition Challenge.
Originally trained for generic object recognition,
they have also proven to be very successful for many other vision tasks \cite{[Razavian14]}.
In this paper, we test the popular
AlexNet \cite{[Krizhevsky12]},
GoogLeNet/Inception \cite{[Szegedy15]},
ResNet \cite{[He16]},
and VGG \cite{[Simonyan14],[Parkhi15]}
architectures.
Features are extracted from different layers, with comparison
between feature vectors done by simple distance or similarity measures.
They are evaluated with a database of frontal periocular images taken with a
digital camera.
Our experiments show that these popular architectures can outperform traditional
features encompassing Local Binary Patterns, Histogram of Oriented Gradients,
and Scale-Invariant Feature Descriptors \cite{[Park11]},
with EER reductions of up to $\sim$40\%.
This shows the effectiveness of those generic objects classifiers to the periocular
recognition problem as well.

Future work includes extending our framework to another periocular databases,
both with visible and near-infrared illumination, and to non-frontal images.
Another approach to cope with the lack of large databases in the periocular domain
is to fine-tune off-the-shelf pre-trained architectures as the ones employed here.
A common approach to reduce complexity is to keep initial layers fixed and only fine-tune some high-level layers,
because earlier layers usually contain generic features useful to many related tasks,
but later layers become progressively more specific to the problem at hand.
Exploiting features from video frames to improve performance is another avenue, thanks
to architectures capable to cope with dynamic information, such as Recurrent Neural Networks.

%

\bibliographystyle{IEEEtran}

\bibliography{fernando1}

\end{document}